\documentclass{article}
\usepackage{spconf,amsmath,graphicx}
\usepackage{colortbl}
\usepackage{amsmath}
\usepackage{hyperref}
\definecolor{Gray}{gray}{0.9}

\title{Activating Frequency and ViT for 3D Point Cloud Quality Assessment without Reference}
\begin{document}
%
\maketitle
\vspace{-3mm}
\begin{abstract}

Deep learning-based quality assessments have significantly enhanced perceptual multimedia quality assessment, however it is still in the early stages for 3D visual data such as 3D point clouds (PCs). Due to the high volume of 3D-PCs, such quantities are frequently compressed for transmission and viewing, which may affect perceived quality. Therefore, we propose no-reference quality metric of a given 3D-PC. Comparing to existing methods that mostly focus on geometry or color aspects, we propose integrating frequency magnitudes as indicator of spatial degradation patterns caused by the compression. To map the input attributes to quality score, we use a light-weight hybrid deep model; combined of Deformable Convolutional Network (DCN) and Vision Transformers (ViT). Experiments are carried out on ICIP20 \cite{perry2020quality}, PointXR \cite{alexiou2020pointxr} dataset, and a new big dataset called BASICS \cite{ak2023basics}. The results show that our approach outperforms state-of-the-art NR-PCQA measures and even some FR-PCQA on PointXR. \textit{The implementation code can be found at:  \href{https://github.com/o-messai/3D-PCQA}{https://github.com/o-messai/3D-PCQA}}
\end{abstract}


\vspace{-3mm}
\section{Introduction}
\label{sec:intro}
\vspace{-3mm}

3D Point clouds (3D-PCs) provide the shape information of 3D objects and can be quickly captured by 3D scanners; which are becoming accessible even in our mobile devices (e.g., tablets, smartphones, etc). Recently, 3D-PC has been an active research field, closely tied to applications such as augmented reality, drones, self-driving vehicles, and 3D video games \cite{qi2018frustum, messai2020adaboost}. However, because of the large number of points cloud required to describe the object, this type of 3D data requires a large amount of memory storage, and demands high computation for transmitting and display. Therefore, implementing compression procedures to the 3D-PCs becomes necessary, which can have an impact on its visual quality. However, to ensure that the compression procedure is reliable, the perceptual quality rate is measured. Subjective and objective studies could be used to get Point Cloud Quality Assessment (PCQA). The former involves human intervention, whereas the latter is based on computational algorithms that anticipate perceptual quality. The PCQA metrics are often used to evaluate the perceptual correctness of the coded point cloud in relation to a certain compression rate. In order to develop sophisticated quality measurements that forecast the perceived impact of a given PC, it is important to validate the metric output with a subjective evaluation, namely human visual quality assessment. It is mostly expressed in terms of Mean Opinion Score (MOS) or Difference Mean Opinion Score (DMOS) \cite{messai20223d, messai2022end}.
Objective quality measurements are classified into three types based on the availability of the reference PC: full-reference (FR), reduced-reference (RR), and no-reference (NR). However, subjective human ratings  can be time consuming and expensive, and a pristine reference is not always accessible. Therefore, researchers are increasingly relying on NR objective measurements due to the broad benefits they give. As a result, most current PCQA approaches are devoted to NR-PCQA in order to meet the criteria of most modern applications. Furthermore, the PCQA metrics can be divided into three categories: point-based metrics, feature-based metrics, and projection-based metrics. In Point-based metrics such as Point-to-Point (Po2Po) \cite{mekuria2016evaluation, tian2017geometric}, Point-to-Plane (Po2Pl) \cite{alexiou2018point}. These metrics work by calculating the geometric/color distance between the reference PC and its distorted variant. In feature-based metrics, geometry and associated properties are extracted at the point level in a global or local way. For instance, a metric called Geotex \cite{diniz2020towards}, which uses Local Binary Pattern (LBP) as descriptors, while in metric \cite{meynet2020pcqm}, both geometry and color data have been used as features. Finally, in projection-based measurements, the points are projected into 2D grids at certain view points/degrees, and 2D quality measures are applied to these views. For instance, in metric \cite{bourbia2022blind}, the authors used point cloud rendering to generate views (2D images), which were then fed into a deep convolutional neural network (CNN) to provide perceptual quality scores.

\begin{figure*}[h]
    \centering
    \includegraphics[height=4.7cm]{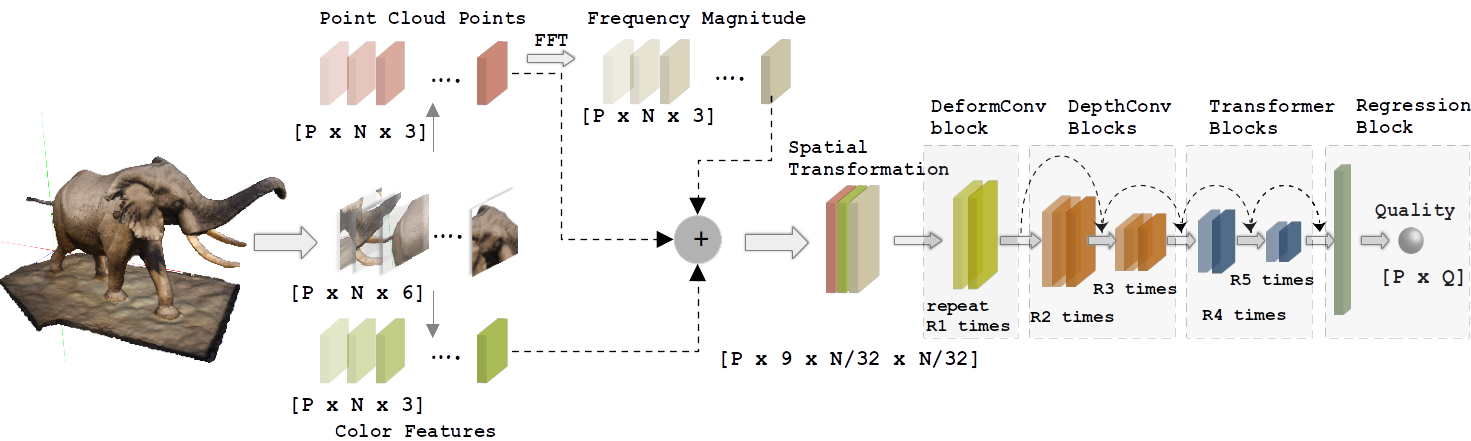}
    \caption{The flowchart of the proposed metric; step 1: extract three input attributes, step 2: map the attributes into a quality score.}
    \label{fig:Features}
\end{figure*}

In the following, we briefly address the recent suggested NR-PCQA metrics: In metric \cite{chetouani2021deep}, low-level details such as geometric distance, local curvature, and brightness values were retrieved from the 3D-PC. The latter inputs were then mapped to a quality score using a deep CNN. Moreover in metric \cite{zhang2022no}, the PCs were transformed from 3D space into domains of quality-related geometry and color features. Then, using 3D natural scene statistics and entropy, features for vector regression (SVR) were extracted. Recently, a new metric \cite{tliba2022representation} based on end-to-end model inspired by Point-Net \cite{qi2017pointnet++}. The model relies on the local inherent properties of subsets of points as inputs without requiring a prior processing steps that could be computationally expensive.
The Graph Convolutional Network (GCN) is increasingly being utilized for processing 3D-PC. For example, in metric \cite{shan2022gpa}, the authors suggested a multi-task GCN model that takes PC inputs and predicts distortion type/degree as well as quality score. Another NR measure proposed \cite{zhang2022mm}, based on deep model that takes both the PC and the related 2D projection as inputs, extracts and aggregates features, and maps them to a quality score. 

Vision Transformer (ViT) \cite{han2022survey} and its derivatives have recently made a significant breakthrough in the field of computer vision problems, showing a greater ability to model global and long-range relationships than CNNs. Since then, transformers have been used for image classification, segmentation, restoration, and so forth. Therefore, in this work, we review the use of ViT for 3D-PC quality assessment, presenting new NR-PCQA. This work makes three contributions, which are stated below:
\vspace{-1mm}
\begin{itemize}
    \item We present a light-weight metric that ranked 1st in term of run-time and 4th in term of accuracy at ICIP 2023 - PCVQA grand challenge (Track 2 \& 4 for no-reference metrics) \cite{ak2023basics}.
    \vspace{-3mm}
    \item We extend the use of ViT for 3D-PC quality assessment using hybrid model; Convolution and self-Attention. We also incorporate deformable convolution since point clouds have non-uniform distributions.
    \vspace{-3mm}
    \item We investigate the use of frequency domain as source of information for assessing the quality of 3D-PC.
\end{itemize}

The remainder of this paper is organized as follows. In Section \ref{sec:Proposed_method}, we describe the proposed method. Then, we present the experimental results in the Section \ref{sec:experimental}. Finally, we give some concluding remarks in Section \ref{sec:conclusion}.


\vspace{-3mm}
\section{PROPOSED METHOD}
\label{sec:Proposed_method}
\vspace{-3mm}
Fig. \ref{fig:Features} presents the flowchart of the proposed system which combines two steps: The first step is to extract effective attributes from the 3D-PC. In the second step, fed the attributes into a deep DCN-ViT model to process and forecast quality. Further details are described in the following subsections. 
\vspace{-5mm}
\subsection{Model input pre-processing}
\vspace{-3mm}
Deep Learning (DL) in particular enabled the automatic extraction of the best high-level features, which outperformed handwritten characteristics. However, learning strategies for Feature-based metrics may change from one measure to the next, but success is strongly dependant on input attributes. Therefore, we create three inputs to be fed into the deep learning model, integrating frequency magnitude, which is new compared to previous measures that rely just on PC color/coordinate data.

For a given 3D-PC, we select $P$ patches with $N$ points as follows:
we initially normalize the PC to a unit sphere and use the furthest point sampling approach to select $M$ number of centroids $(i.e., C = M_{1}, M_{2}, M_{3},..., M_{p})$. Then, we use the K-nearest neighbors clustering algorithm to build a patch around each centroid. In our experiments, $P$ was set to $100$, where $K=N$ was set to 1024. Afterwards, as shown in Fig. \ref{fig:Features}, RGB color information, point coordinates $(i.e., x, y, z)$ and its frequency magnitude are provided for each patch. Resulting tensor of size $[P \times N \times 3]$ for each input attribute.

\vspace{-3mm}
\subsection{Frequency Magnitude of 3D point Clouds}
\begin{figure}[t]
    \centering
    \includegraphics[height=3.0cm]{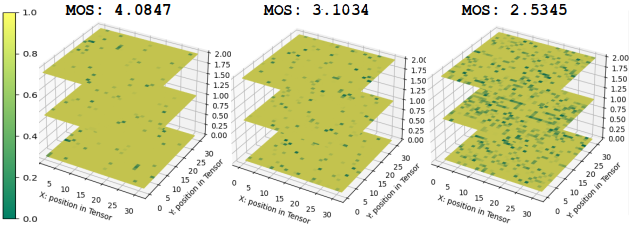}
    \caption{Examples of frequency magnitudes of different quality PC, (values have been normalized to $[0-1]$ scale for better visualization). Three input tensors of size $[3 \times 32 \times 32]$}
    \label{fig:freq}
\end{figure}

\vspace{-3mm}
Obtaining the frequency magnitude of the 3D-PC could be relevant in some applications. The frequency magnitude is a measure of the spatial frequency content of the point cloud data, and it can provide information about the overall shape and structure of the object represented by the point cloud, as well as an indicator of spatial patterns caused by degradation/compression. 
In this work, the magnitude is computed using the Fast Fourier Transform (FFT) on the extracted patch of point coordinates (e.i., $[P \times N \times 3]$ ), the magnitude is then calculated by taking the absolute value of the Fourier coefficients. Finally, we simplify the interpretation for the deep learning model by relocating the zero frequency component to the center of the spectrum and also the RGB data is standardized to a $[0-1]$ scale. The Fourier transform is applied to each spatial dimension of the 3D point cloud coordinates, producing an output tensor of the same shape as the input. Fig. \ref{fig:freq}. shows an examples of different quality of the same PC patch reshaped to size of $[3 \times \frac{N}{32}\times \frac{N}{32}]$. As can be seen, the frequency magnitudes differ at different MOS quality scores, with lower quality scores (e.i., $MOS: 2.5345$) corresponding to low-frequency in the 3D-PC data and high-density regions corresponding to high-frequency. This investigation inspired us to include frequency as an indicator characteristic in our system.


\vspace{-3mm}
\subsection{Spatial Transformation}
\vspace{-3mm}
Depending on the task and model architecture, reshaping the input data into an image-like shape might provide numerous benefits to the deep learning model. By reshaping features into an image-like shape, we take advantage of the convolutional layers of CNNs to perform local feature extraction and capture spatial relationships between features. Therefore, we concatenate and reshape the three input attributes into one tensor of size $[P \times 9 \times \frac{N}{32} \times \frac{N}{32}]$.
However, converting the point cloud to an image-like format may result in the loss of certain information about the point cloud, such as the point ordering and local neighborhood connections. This may prevent the model's ability to learn correctly from point cloud data. As a result, we use a deformable convolutional block as first layer in our model to help it learn to account for the non-uniform distribution of 3D-PCs.

\vspace{-3mm}
\subsection{Deep CNN-ViT model}
\vspace{-3mm}

We present a hybrid model architecture based on deformable convolution, depth-wise convolution, and ViT, inspired by the work of \cite{dai2021coatnet}. The benefits of the stated techniques addressed in the following: 
The deformable convolution makes the receptive field flexible and can be adjusted to the distribution of point cloud data. This means that the convolutional filter may learn a more accurate representation of the underlying features, resulting in higher accuracy in the quality prediction task. Depth-wise convolution applies a distinct filter to each channel of a given input. Unlike classical convolution, which uses a single filter across all channels, depth-wise convolution uses a distinct filter for each channel, thereby learning independent feature maps and thus requiring fewer parameters to learn. While the convolution extracts local feature maps. The self-attention mechanism in ViT extracts global feature maps.

As illustrated in Fig.\ref{fig:Features}, our model is build up of the following blocks in the correct order, with skip connections and concatenations between them: 
1 $DeformConv$ block, followed by 2 $DepthConv$ blocks and 2 $Transformer$ blocks. The extracted feature maps from the previous blocks are then input into the $Regression$ block, involving global average pooling followed by a Fully Connected (FC) layer of output $[1 \times 1]$ for quality score. As suggested in model \cite{dai2021coatnet}, we always use the kernel size 3 for $DeformConv$ and $DepthConv$ blocks. While for the $Transformer$ block, we set the size of each attention head to 32. For the repeat block parameters, were set as: $R1=3,R2=3, R3=6, R4=14,$ and $R5=2$. The corresponding size of hidden channels respectively are: $D1=64,D2=96, D3=128, D4=128,$ and $D5=512$. However, more implementation details, can be found in the source code. The model processes $P$ patches from each 3D-PC, and the anticipated quality score $Q_{f}$ of the entire PC is presented by computing the mean of patch scores: $Q_{f} = \frac{1}{P} \sum Q_{p}$.


\vspace{-3mm}
\subsection{Model training}
\vspace{-3mm}
In order to minimize the error during training of the designed model, we use the $SmoothL1Loss$ function which is a variant of the $L_{1}$ loss functions as described in the following:
\vspace{-2mm}
\begin{multline}
     \text{SmoothL1Loss}(MOS, Q) = \\
    \frac{1}{p} \sum_{i=1}^{p} \begin{cases}
    0.5 (MOS - Q)^2, & \text{if } |MOS - Q| < 1 \\
    |MOS - Q| - 0.5, & \text{otherwise}
  \end{cases}   
\end{multline}
where $Q$ are the predicted quality. The $MOS$, refers to the human rating of the 3D-PC, and $n$ is the number of patches. $SmoothL1Loss$ reduces the sensitivity to outliers by smoothing the loss function near zero. It accomplishes this by employing a piecewise function that for large errors behaves like $L1 loss$ but is quadratic (e.i., $L2 loss$) for small errors. This makes it more appropriate for our MOS regression task, where the data may contain outliers where a more robust loss function is required.
To update the weights of the model (8 million parameters), we used the Stochastic Gradient Descent (SGD) with a momentum factor equals to 0.9, a weight decay factor sets to $10^{-4}$, a mini batch size equals to 128 and a learning rate initialized to $10^{-5}$. The Pytorch framework was used to implement our approach.
\vspace{-3mm}

\vspace{-3mm}
\section{EXPERIMENTAL RESULTS}
\label{sec:experimental}
\vspace{-3mm}
\subsection{Dataset and training protocol}
\vspace{-3mm}
A database of 3D-PCs with quality scores for PC is required for training and evaluating. The quality score is often obtained by MOS subjective scoring. For the PCQA domain, a variety of databases are publicly available. Three datasets were used for performance evaluation: two well-known (PointXR \cite{alexiou2020pointxr} and ICIP \cite{perry2020quality}) and a third new dataset named BASICS \cite{ak2023basics}, which was part of the ICIP 2023 Grand Challenge on Point Cloud Quality Assessment (PCVQA). The datasets are briefly described as follows: PointXR contains 5 PCs, from which 45 degraded versions were created using G-PCC with octree coding for geometry compression and Lifting and RAHT for color compression. ICIP20 contains 6 reference PCs from which 90 degraded copies were created using 3 different compression methods: V-PCC, G-PCC with triangle soup coding, and G-PCC with octree coding. Each reference PC was compressed at 5 distinct levels. BASICS, the largest dataset currently available, contains 75 references. Each point cloud was compressed using four distinct compression methods with distinct compression levels (VPCC, GPCC-RAHT, GPCC-Predlift, and GEOCNN), yielding 1494 processed PC.
The performance of our method was quantified using the three databases. To guarantee that our model evaluates the PC quality rather than focusing on the content, we divide each database into 80\% for training and rest 20\% for test based on PC references. So the training PCs data are independent from those used in test phase. We repeat the same process 5 times and report the average performance. During the training, we do not consider data augmentation since at each training epoch, a random patches were cropped. Over 500 training epochs, the best model was chosen for test. 

\vspace{-5mm}
\subsection{Comparison with the State-of-the-Art}
\vspace{-3mm}
The performance has been measured across three metrics: The \emph{RMSE}, Pearson linear correlation coefficient (\emph{PLCC}), Spearman's rank order correlation coefficient (\emph{SROCC}) between the machine quality judgments (objective scores) and the human ratings (subjective scores). High values for \emph{PLCC} and \emph{SROCC} (close to 1) and low values for \emph{RMSE} (close to 0) indicate a better prediction performance. 
Overall, the statistical association between human quality scores and our method ratings exhibited good performance and consistency. The obtained results were compared to many FR and NR-PCQA. Among them, two are recent reference-free metrics based on the use of CNN models, namely Tliba \cite{tliba2022representation} and GQI-VGG19 \cite{chetouani2021deep}. Table \ref{T:results-all} shows the results of these methods on both ICIP20 and PointXR datasets. Best metric is represented on bold. As can be seen, our metric outperforms all the state-of-the-art NR and FR metrics on PointXR, but on ICIP20 and BASICS the performance falls short.
These findings are supported by the fact that the two datasets are built of a more diverse collection of compression techniques, whereas PointXR comprises only G-PCC method with octree coding for geometry compression and Lifting and RAHT for color compression. Furthermore, we report the performance of our method according to the size of the training set. Table \ref{T:DP} shows the correlations achieved for a training set of size 50\%, 70\% and 80\%. The partition ratio has a slight impact on the performance. And it does not suffer from an over-fitting problem. 

\vspace{-5mm}
\begin{table}[ht]
\centering
\caption{Overall performance comparison on ICIP20, PointXR and BASICS datasets.}
\resizebox{1\columnwidth}{!}{%
\begin{tabular}{c|c|c c|| c c|| c c} 

&  &  \textbf{ICIP20}                &   & \textbf{PointXR}           &   & \textbf{BASICS}           & \\ \hline
Type          & Metrics & SROCC        &  PLCC    & SROCC        &  PLCC    & SROCC        &  PLCC \\  \hline 

&Po2pointMSE         &0.950 &0.945 &0.978 &0.887 &- &- \\   

FR &Po2planeMSE       &0.959 &0.945 &0.942 &0.855 &- &-\\   

&PSNRpo2pointMSE         &0.934 &0.880 &0.978 &0.983  &- &-\\   

&PSNRpo2planeMSE         &0.953 &0.916 &0.950 &0.972 &- &-\\ 
\hline  

&Tliba \cite{tliba2022representation}         &0.955 &0.908 &0.970 &0.964 &- &-\\   

 NR &GQI-VGG19 \cite{chetouani2021deep}        &\textbf{0.966} &\textbf{0.952} &- &-&- &- \\   

 
\rowcolor{Gray} &  Proposed   &0.893 &0.849 &\textbf{0.988} &\textbf{0.981}  &\textbf{0.710} &\textbf{0.764} \\ \hline 
\end{tabular}
\label{T:results-all}
}
\end{table}

\vspace{-1mm}

\begin{table}[ht]
\centering
\caption{Performance of the proposed metric under different train-test partitions on PointXR.}
\resizebox{0.6\columnwidth}{!}{%
\begin{tabular}{c|c c c}

\hline
Partition & SROCC      &  PLCC    & RMSE\\  
\hline 

80\%-20\% &\textbf{0.988} &\textbf{0.981}  &\textbf{1.327}   \\ 

70\%-30\%  &0.915 &0.942  &1.378   \\ 

50\%-50\%   &0.884 &0.921  &1.449  \\ 

\hline 

\end{tabular}
} 
\label{T:DP}
\end{table}

\subsection{Ablation study and run-time}
\vspace{-3mm}
We simply delete one of the extracted attributes from the input data in the ablation test scenario. As a result, following the same experiment protocol, the model is completely retrained and tested. Table \ref{T:ablation_study} compares performance without and with frequency magnitude information. In addition, we studied the use of RGB information data. The results indicate that our approach improves performance and supports the idea of using frequency of PC data as a visual quality attribute. As expected, involving the RGB color information of the PC improves the performance since it is necessary to describe the 3D object.
\vspace{-2mm}
\begin{table}[htbp]
\centering
\caption{Performance obtained of ablation tests on PointXR.}
\resizebox{0.7\columnwidth}{!}{%
\begin{tabular}{c|c c c}
\hline
Model  & SROCC        &  PLCC    & RMSE\\  
\hline 

Without RGB data &0.962 &0.952  &1.642  \\ 

Without Frequency &0.976 &0.962  &1.403 \\ 

\textbf{Proposed}  &\textbf{0.988} &\textbf{0.981}  &\textbf{1.327}\\ 
   
\hline 
\end{tabular}
} 
\label{T:ablation_study}
\end{table}
\vspace{-2mm}
We measured less than 2 seconds of run-time (\textbf{1.129 s}) using a single 3D-PC, including patches extraction, processing and loading using a Dell Precision 5570 laptop equipped with an Intel i7-12800H CPU @ 4.80GHz processor and an NVIDIA Quadro RTX A 2000 GPU. In terms of run-time speed, the proposed metric has the potential to be used for real-time applications.

\vspace{-5mm}
\section{CONCLUSION}
\label{sec:conclusion}
\vspace{-3mm}
In this paper, we introduced a new NR-PCQA approach based on a hybrid model (DCN-ViT) combining deformable convolution and self-attention for evaluating the quality of given 3D-PC. The suggested light-weight model takes RGB color information, point coordinates, and frequency magnitude of the PC as inputs to determine the visual quality. The ablation study demonstrates that frequency analysis is beneficial and has potential for further work. Based on a comparative examination using three datasets, our model outperforms state-of-the-art approaches on PointXR and competitive results on ICIP20 and BASICS dataset. As future work, we will undertake in-depth examination and consider extending the model to handle auxiliary tasks such as: distortion type recognition and degree of deformations.
\vspace{-5mm}

\bibliographystyle{IEEEbib}
\bibliography{strings,refs}

\begin{thebibliography}{10}

\bibitem{perry2020quality}
Stuart Perry, Huy~Phi Cong, Lu{\'\i}s~A da~Silva~Cruz, Jo{\~a}o Prazeres,
  Manuela Pereira, Antonio Pinheiro, Emil Dumic, Evangelos Alexiou, and Touradj
  Ebrahimi,
\newblock ``Quality evaluation of static point clouds encoded using mpeg
  codecs,''
\newblock in {\em 2020 IEEE International Conference on Image Processing
  (ICIP)}. IEEE, 2020, pp. 3428--3432.

\bibitem{alexiou2020pointxr}
Evangelos Alexiou, Nanyang Yang, and Touradj Ebrahimi,
\newblock ``Pointxr: A toolbox for visualization and subjective evaluation of
  point clouds in virtual reality,''
\newblock in {\em 2020 Twelfth International Conference on Quality of
  Multimedia Experience (QoMEX)}. IEEE, 2020, pp. 1--6.

\bibitem{ak2023basics}
Ali Ak, Emin Zerman, Maurice Quach, Aladine Chetouani, Aljosa Smolic, Giuseppe
  Valenzise, and Patrick~Le Callet,
\newblock ``Basics: Broad quality assessment of static point clouds in
  compression scenarios,''
\newblock {\em arXiv preprint arXiv:2302.04796}, 2023.

\bibitem{qi2018frustum}
Charles~R Qi, Wei Liu, Chenxia Wu, Hao Su, and Leonidas~J Guibas,
\newblock ``Frustum pointnets for 3d object detection from rgb-d data,''
\newblock in {\em Proceedings of the IEEE conference on computer vision and
  pattern recognition}, 2018, pp. 918--927.

\bibitem{messai2020adaboost}
Oussama Messai, Fella Hachouf, and Zianou~Ahmed Seghir,
\newblock ``Adaboost neural network and cyclopean view for no-reference
  stereoscopic image quality assessment,''
\newblock {\em Signal Processing: Image Communication}, vol. 82, pp. 115772,
  2020.

\bibitem{messai20223d}
Oussama Messai, Aladine Chetouani, Fella Hachouf, and Zianou~Ahmed Seghir,
\newblock ``3d saliency guided deep quality predictor for no-reference
  stereoscopic images,''
\newblock {\em Neurocomputing}, 2022.

\bibitem{messai2022end}
Oussama Messai and Chetouani,
\newblock ``End-to-end deep multi-score model for no-reference stereoscopic
  image quality assessment,''
\newblock in {\em 2022 IEEE International Conference on Image Processing
  (ICIP)}. IEEE, 2022, pp. 2721--2725.

\bibitem{mekuria2016evaluation}
RN~Mekuria, Zhu Li, C~Tulvan, and P~Chou,
\newblock ``Evaluation criteria for pcc (point cloud compression),''
\newblock 2016.

\bibitem{tian2017geometric}
Dong Tian, Hideaki Ochimizu, Chen Feng, Robert Cohen, and Anthony Vetro,
\newblock ``Geometric distortion metrics for point cloud compression,''
\newblock in {\em 2017 IEEE International Conference on Image Processing
  (ICIP)}. IEEE, 2017, pp. 3460--3464.

\bibitem{alexiou2018point}
Evangelos Alexiou and Touradj Ebrahimi,
\newblock ``Point cloud quality assessment metric based on angular
  similarity,''
\newblock in {\em 2018 IEEE International Conference on Multimedia and Expo
  (ICME)}. IEEE, 2018, pp. 1--6.

\bibitem{diniz2020towards}
Rafael Diniz, Pedro~Garcia Freitas, and Myl{\`e}ne~CQ Farias,
\newblock ``Towards a point cloud quality assessment model using local binary
  patterns,''
\newblock in {\em 2020 Twelfth International Conference on Quality of
  Multimedia Experience (QoMEX)}. IEEE, 2020, pp. 1--6.

\bibitem{meynet2020pcqm}
Gabriel Meynet, Yana Nehm{\'e}, Julie Digne, and Guillaume Lavou{\'e},
\newblock ``Pcqm: A full-reference quality metric for colored 3d point
  clouds,''
\newblock in {\em 2020 Twelfth International Conference on Quality of
  Multimedia Experience (QoMEX)}. IEEE, 2020, pp. 1--6.

\bibitem{bourbia2022blind}
Salima Bourbia, Ayoub Karine, Aladine Chetouani, and Mohammed El~Hassouni,
\newblock ``Blind projection-based 3d point cloud quality assessment method
  using a convolutional neural network.,''
\newblock in {\em VISIGRAPP (4: VISAPP)}, 2022, pp. 518--525.

\bibitem{chetouani2021deep}
Aladine Chetouani, Maurice Quach, Giuseppe Valenzise, and Fr{\'e}d{\'e}ric
  Dufaux,
\newblock ``Deep learning-based quality assessment of 3d point clouds without
  reference,''
\newblock in {\em 2021 IEEE International Conference on Multimedia \& Expo
  Workshops (ICMEW)}. IEEE, 2021, pp. 1--6.

\bibitem{zhang2022no}
Zicheng Zhang, Wei Sun, Xiongkuo Min, Tao Wang, Wei Lu, and Guangtao Zhai,
\newblock ``No-reference quality assessment for 3d colored point cloud and mesh
  models,''
\newblock {\em IEEE Transactions on Circuits and Systems for Video Technology},
  vol. 32, no. 11, pp. 7618--7631, 2022.

\bibitem{tliba2022representation}
Marouane Tliba, Aladine Chetouani, Giuseppe Valenzise, and Fr{\'e}d{\'e}ric
  Dufaux,
\newblock ``Representation learning optimization for 3d point cloud quality
  assessment without reference,''
\newblock in {\em 2022 IEEE International Conference on Image Processing
  (ICIP)}. IEEE, 2022, pp. 3702--3706.

\bibitem{qi2017pointnet++}
Charles~Ruizhongtai Qi, Li~Yi, Hao Su, and Leonidas~J Guibas,
\newblock ``Pointnet++: Deep hierarchical feature learning on point sets in a
  metric space,''
\newblock {\em Advances in neural information processing systems}, vol. 30,
  2017.

\bibitem{shan2022gpa}
Ziyu Shan, Qi~Yang, Rui Ye, Yujie Zhang, Yiling Xu, Xiaozhong Xu, and Shan Liu,
\newblock ``Gpa-net: No-reference point cloud quality assessment with
  multi-task graph convolutional network,''
\newblock {\em arXiv preprint arXiv:2210.16478}, 2022.

\bibitem{zhang2022mm}
Zicheng Zhang, Wei Sun, Xiongkuo Min, Quan Zhou, Jun He, Qiyuan Wang, and
  Guangtao Zhai,
\newblock ``Mm-pcqa: Multi-modal learning for no-reference point cloud quality
  assessment,''
\newblock {\em arXiv preprint arXiv:2209.00244}, 2022.

\bibitem{han2022survey}
Kai Han, Yunhe Wang, Hanting Chen, Xinghao Chen, Jianyuan Guo, Zhenhua Liu,
  Yehui Tang, An~Xiao, Chunjing Xu, Yixing Xu, et~al.,
\newblock ``A survey on vision transformer,''
\newblock {\em IEEE transactions on pattern analysis and machine intelligence},
  vol. 45, no. 1, pp. 87--110, 2022.

\bibitem{dai2021coatnet}
Zihang Dai, Hanxiao Liu, Quoc~V Le, and Mingxing Tan,
\newblock ``Coatnet: Marrying convolution and attention for all data sizes,''
\newblock {\em Advances in Neural Information Processing Systems}, vol. 34, pp.
  3965--3977, 2021.

\end{thebibliography}

\end{document}